\documentclass[letterpaper, 10 pt, conference]{ieeeconf}  

\IEEEoverridecommandlockouts                              

\usepackage{cite}
\usepackage{amsmath,amssymb,amsfonts}
\usepackage{multirow}
\usepackage{graphicx}
\usepackage{float}
\usepackage{subfigure}
\usepackage[linesnumbered,ruled,vlined]{algorithm2e}
\usepackage{booktabs}
\usepackage{array}
\usepackage{tikz}
\usepackage{pgfplots}
\usepackage{comment}

\pgfmathdeclarefunction{gauss}{2}{%
  \pgfmathparse{1/(#2*sqrt(2*pi))*exp(-((x-#1)^2)/(2*#2^2))}%
}

\title{\LARGE \bf
PathRL: An End-to-End Path Generation Method for Collision Avoidance via Deep Reinforcement Learning
}

\author{Wenhao Yu$^{1}$, Jie Peng$^{2}$, Quecheng Qiu$^{3}$, Hanyu Wang$^{2}$, Lu Zhang$^{4}$ and Jianmin Ji$^{3, *}$
\thanks{$^{1}$ Institute of Advanced Technology, University of Science and Technology of China (USTC), Hefei 230026, China
        {\tt\small wenhaoyu@mail.ustc.edu.cn}}%
\thanks{$^{2}$ School of Data Science, USTC, Hefei 230026, China
        {\tt\small \{pengjieb, wangyuu\}@mail.ustc.edu.cn}}%
\thanks{$^{3}$ School of Computer Science and Technology, USTC, Hefei 230026, China
        {\tt\small {qiuqc}@mail.ustc.edu.cn}}%
\thanks{$^{4}$ Institute of Artificial Intelligence, Hefei Comprehensive National Science Center, 230088, China
        {\tt\small {luzha}@ustc.edu.cn}}%
\thanks{${*}$ Corresponding author. {\tt\small jianmin@ustc.edu.cn}}
}

\begin{document}

\maketitle
\thispagestyle{empty}
\pagestyle{empty}


\begin{abstract}
    Robot navigation using deep reinforcement learning (DRL) has shown great potential in improving the performance of mobile robots.
    Nevertheless, most existing DRL-based navigation methods primarily focus on training a policy that directly commands the robot  with low-level controls, like linear and angular velocities, which leads to unstable speeds and unsmooth trajectories of the robot during the long-term execution.
    An alternative method is to train a DRL policy that outputs the navigation path directly. Then the robot can follow the generated path smoothly using sophisticated velocity-planning and path-following controllers, whose parameters are specified according to the hardware platform. 
    However, two roadblocks arise for training a DRL policy that outputs paths:
    (1) The action space for potential paths often involves higher dimensions comparing to low-level commands, which increases the difficulties of training;
    (2) It takes multiple time steps to track a path instead of a single time step, which requires the path to predicate the interactions of the robot w.r.t. the dynamic environment in multiple time steps. This, in turn, amplifies the challenges associated with training.
    In response to these challenges, we propose \textit{PathRL}, a novel DRL method that trains the policy to generate the navigation path for the robot. 
    Specifically, we employ specific action space discretization techniques and tailored state space representation methods to address the associated challenges. Curriculum learning is employed to expedite the training process, while the reward function also takes into account the smooth transition between adjacent paths.
   In our experiments, \textit{PathRL} achieves better success rates and reduces angular rotation variability 
   compared to other DRL navigation methods, facilitating stable and smooth robot movement.
    We demonstrate the competitive edge of \textit{PathRL} in both real-world scenarios and multiple challenging simulation environments.
\end{abstract}

\section{Introduction}

DRL navigation methods have shown great potential in improving the flexibility and adaptability of mobile robots in complex and changing scenarios~\cite{xiaoMotionPlanningControl2022a}. 
Most existing DRL navigation methods intend to train a policy that directly commands the robot with single-step low-level controls, like linear and angular velocities.
However, it is challenging for most DRL algorithms to memorize long histories of low-level controls, which may lead to unstable speeds and unsmooth trajectories of the robot during the long-term execution and cannot achieve typical driving maneuvers~\cite{wang2023efficient}.
It is also hard to train a DRL policy converging to the almost zero exploration from the balance of \textit{exploration and exploitation}~\cite{coggan2004exploration}, which makes the outputs of the policy easy to drift.
On the other hand, a human-friendly mobile robot is expected to drive stably and smoothly.

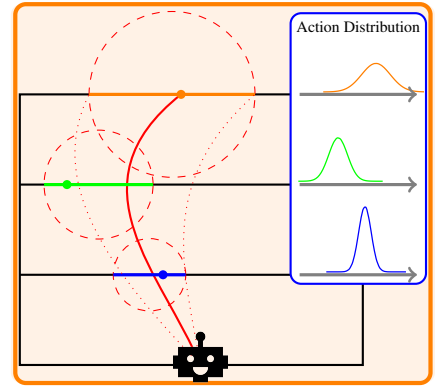
\begin{figure}[t]
\setlength{\abovecaptionskip}{-0.5em}
\centering
\begin{tikzpicture}[scale=0.6]
	\fill[rounded corners, orange, ultra thick, fill opacity=0.1] (-5.2, -0.4) rectangle (4.2, 8.0);
	\draw[rounded corners, orange, ultra thick] (-5.1, -0.4) rectangle (4.1, 8.0);
	\draw[thick] (-5, 0) rectangle (2.6, 6);
	\draw[thick] (-5, 2) -- (2.6, 2);
	\draw[thick] (-5, 4) -- (2.6, 4);
	\fill[rounded corners, white, thick] (1.0, 1.8) rectangle (4.0, 7.8);
	\draw[rounded corners, blue, thick] (1.0, 1.8) rectangle (4.0, 7.8);
	\draw[red, thick] (-1, 0) .. controls (-1.826845, 2) and (-3.95503, 4) .. (-1.426845, 6);
	\draw[red, dashed] (-2.126845, 2) circle (0.80256);
        \draw[blue, very thick] (-2.9294, 2) -- (-1.3243, 2);
	\draw[red, dashed] (-3.25503, 4) circle (1.20741);
        \draw[green, very thick] (-4.46244, 4) -- (-2.04762, 4);
	\draw[red, dashed] (-1.626845, 6) circle (1.83807);
        \draw[orange, very thick] (-3.464915, 6) -- (0.211225, 6);
	\fill[blue] (-1.826845, 2) circle (.1cm);
	\fill[green] (-3.95503, 4) circle (.1cm);
	\fill[orange] (-1.426845, 6) circle (.1cm); 
	\draw[red, dotted] (-1, 0) .. controls (-1.324285, 2) and (-2.04762, 4) .. (0.211225, 6);
	\draw[red, dotted] (-1, 0) .. controls (-2.929405, 2) and (-4.46244, 4) .. (-3.464915, 6);	
	\begin{axis}[style={mark=none, samples=100, smooth}, clip=false, hide axis]
	\addplot[domain=-2.1:1.3, blue, thick, xscale=0.4, yscale=0.3, xshift=2.4cm, yshift=5.3cm] {gauss(-0.450738,0.26752)};
	\addplot[domain=-2.6:1.0, green, thick, xscale=0.4, yscale=0.3, xshift=1.5cm, yshift=12.0cm] {gauss(-0.902012,0.40247)};
	\addplot[domain=-2.5:1.8, orange, thick, xscale=0.4, yscale=0.3, xshift=2.75cm, yshift=18.6cm] {gauss(-0.250738,0.61269)};
	\end{axis}
	
	\draw[gray, very thick, <-] (3.8, 2.0) -- (1.2, 2.0);
	\draw[gray, very thick, <-] (3.8, 4.0) -- (1.2, 4.0);
	\draw[gray, very thick, <-] (3.8, 6.0) -- (1.2, 6.0);
	
	\node[ultra thick, font=\fontsize {6} {8}\selectfont] at (2.5,7.5) {Action Distribution};
	\draw (-1.0,0.18) node {\includegraphics[scale=0.05]{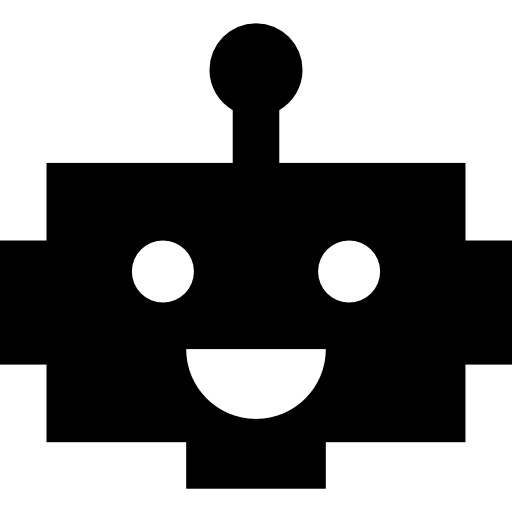}};	
\end{tikzpicture}
\caption{Illustration of the path generation. The area surrounded by two red dotted lines denotes the potential path area. The control points in different colors are sampled by the DRL policy, and the final sampling path (red solid line) is formed through path continuation (B\'{e}zier curve).}
\label{fig:first}
\vspace{-2.0em}
\end{figure}

The decoupling of the control module from the planning module enables the robot to \textit{follow fluctuating path outputs using stable control actions.}
In this paper, we propose \textit{PathRL}, a DRL-based navigation method that outputs navigation paths, which can be followed smoothly by the robot using sophisticated path-tracking algorithms~\cite{coulter1992implementation, hellstrom2006follow, koh1999smooth}.
This form of DRL network makes the robot action predictable due to the low-level controls that are got by the control module, which is predictable in multiple time steps.
Moreover, we also benefit from the flexibility and adaptability of the DRL policy.
The behavior policy limits the path generated by the imitation learning-based method, and our proposed \textit{PathRL} as a policy-based DRL algorithm employs more flexibility.
In \textit{PathRL}, the trained policy focuses on generating proper paths for the robot in various dynamic environments, which is more robust for various robots with different motion models and can also benefit human-robot interaction~\cite{preece1994human,mackenzie2012human}. 



It is challenging to train a navigation policy that outputs paths as the paths often involve the higher-dimensional policy space.
As shown in Fig.~\ref{fig:first}, in \textit{PathRL}, we use B\'{e}zier curve to fit the navigation path via the current position of the robot and other three control points generated by the DRL policy.
Notice that, each pair of adjacent control points has a fixed interval in the longitudinal direction. 
Then these three control points can be specified by the DRL policy via three continuous numbers for corresponding values in the horizontal direction. To facilitate subsequent citation and description, we name the above method of setting control point coordinates as semi-discrete operations.
In this way, the DRL policy can specify a path via these three continuous numbers, which is slightly more complex than the DRL method in~\cite{chen2020distributed} where the policy outputs two continuous numbers corresponding to linear and angular velocities, respectively.
In \textit{PathRL}, we employ three consecutive path-based frames of sensor information and relative targets as inputs of the navigation policy, as shown in Fig.~\ref{fig:overview}. The state of each frame is extracted from the state space upon the termination of semi-MDP options~\cite{sutton1999between} (once a path has been completely followed). We also apply curriculum learning~\cite{bengio2009curriculum} to speed up the training process.

In our experiments, we employ the soft actor-critic (SAC) algorithm~\cite{haarnoja2018soft} as the DRL method for the training.
During the training, the robot is driven by a path-following algorithm that follows the paths generated by the trained policy while given a fixed linear velocity. 
The robot will follow the new path only after it has completed the following of the previous one, which follows a typical semi-MDP process~\cite{sutton1999between} with temporal abstraction~\cite{merel2018neural, merel2020catch}. 
During the execution, this path-following process can be interrupted when the robot is approaching obstacles to achieve more safety navigation.
Our experiments in both real-world scenarios and multiple challenging simulation environments show that  \textit{PathRL} enables an explicable and predictable DRL navigation policy while maintaining the strategic diversity of DRL agents.

The main contributions of our work are summarized in the following:
\begin{itemize}

    \item We propose a novel end-to-end DRL-based method, \textit{PathRL}, that directly outputs navigation paths without relying on the supervised learning paradigm and is competent for a variety of complex scenarios.
    \item We implement semi-discrete operations on the action space to reduce the difficulty of policy training. We utilize three consecutive path-based frames as network inputs to enhance the performance of navigation strategies in dynamic and complex environments. We design a reward function to ensure that the output path is smoother and satisfies the smooth transition between adjacent paths. 
    \item Extensive experiments demonstrate the promise of the \textit{PathRL} in achieving superior yet smoother driving trajectories.
\end{itemize}

\section{Related work}

Classical navigation methods, like Time Elastic Band (TEB) algorithm~\cite{rosmann2012trajectory}, Optimal Reciprocal Collision Avoidance (ORCA)~\cite{berg2011reciprocal} and Social Force Model (SFM)~\cite{helbing1995social}, may take a lot of hard work to find proper parameters in various scenarios.
Supervised learning based navigation methods, like \cite{aminiVariationalEndtoEndNavigation2019} and \cite{maNavigationPreciseLocalization2019}, rely heavily on a large amount of labeled data, while the quality of which directly affects the performance of the trained policy. 


Recently, there have been many DRL-based works on robot motion planning and collision avoidance.
\cite{tai2017virtual} used the asynchronous deep deterministic policy gradients (ADDPG) algorithm with low-dimensional LiDAR data (10 returns) as input to create a navigation system with good obstacle avoidance capabilities. In the field of social navigation, \cite{yao2021crowd} provided a crowd navigation method where both the pedestrian map and the sensor map were used to allow pedestrians to follow different collision avoidance strategies. In these works, the DRL network directly replaces the entire navigation stack and outputs end-to-end navigation instructions.
However, these end-to-end methods try to predict low-level control commands for the robot, which may lead to unstable actions and unsmooth driving trajectories. 

APF-RL~\cite{bektacs2022apf} combined the strengths of Artificial Potential Functions (APF) with DRL, employing the SAC algorithm to dynamically adjust the two input parameters of the APF controller: the intermediate goal and the k-parameter. PRM-RL~\cite{faust2018prm} used PRM as a global planner and DRL as a local planner to complete long-range navigation tasks. The majority of these methods incorporate elements of traditional approaches, with some serving as learning subsystems within classic architectures, while others focus on training individual components of traditional methods. The emphasis here lies in highlighting the distinction of our approach from previous methods. We aim to leverage the flexibility and adaptability inherent in DRL to train an end-to-end policy network capable of directly generating paths. 

The following works are more similar to our method in the idea of path generation. \cite{wang2023efficient} leveraged expert prior knowledge to learn high-level motion skills instead of low-level control skills.
In contrast, our method focuses on searching feasible paths directly from the policy space in an end-to-end manner without any expert prior knowledge.
\cite{zhang2022robot} addressed the problem of predicting proper paths for robots. In this work, the navigation problem is modeled as a deep Markov model, the driving distance and rotation angle of the robot are output at each step, and the final navigation path is formed by combining the multi-step outputs of the algorithm. The method in~\cite{zhang2022robot} can achieve good performance in static scenarios, however, it does not perform well in dynamic scenarios.

\section{Method}

\subsection{Reinforcement Learning Components}
In DRL, we formulate robot's navigation as a Markov Decision Process (MDP) problem. Specifically, an MDP is a tuple $\{\mathcal{S}, \mathcal{A}, \mathcal{P}, \mathcal{R}, \gamma\}$, where $\mathcal{S}$ is the state space, $\mathcal{A}$ is the action space, $\mathcal{P}$ presents the transition probability between states, and $\gamma$ is the discount factor in $(0,1)$.

\subsubsection{State Space} \label{method: observation}
In our formulation, a state $s_t=(m_t, g_t)$ at time step $t$ consists of two parts: an egocentric costmap $m_t$ and a relative target pose $g_t$. 

The egocentric costmap\footnote{http://wiki.ros.org/costmap\_2d} is a $84\times84$ grid map that is generated by a 3D laser sensor with $180^{\circ}$ field of view and presents information about the environment around the robot, including the shape of the robot and the observable appearances of various obstacles.

The relative target pose $g_t = (x^g_t, y^g_t, \alpha^g_t)$ of the robot includes the relative position of the navigation target point $(x^g_t, y^g_t)$ and the relative target orientation $\alpha^g_t$  of the robot w.r.t. the current global pose $(x_t, y_t, \alpha_t)$ of the robot at time $t$, i.e., this relative target pose $g_t$ is the local pose for the local coordinate system when the pose of the robot is identified as the origin. 

In pathRL, the input of the DRL network is three consecutive path-based frames.
These three frames contain the sensor information and relative target poses of the robot at the poses of $(x_t^{-2}, y_t^{-2}, \alpha_t^{-2})$, $(x_t^{-1}, y_t^{-1}, \alpha_t^{-1})$, and $(x_t, y_t, \alpha_t)$, respectively. 
Notice that, we define the two nearest consecutive paths that the robot must follow to reach its current pose $(x_t, y_t, \alpha_t)$ as $P^{-1}_t$ and $P^{-2}_t$, respectively. We use $(x_t^{-1}, y_t^{-1}, \alpha_t^{-1})$ and $(x_t^{-2}, y_t^{-2}, \alpha_t^{-2})$ to denote the starting poses of the robot for these two paths, respectively. 

\begin{figure}[ht]
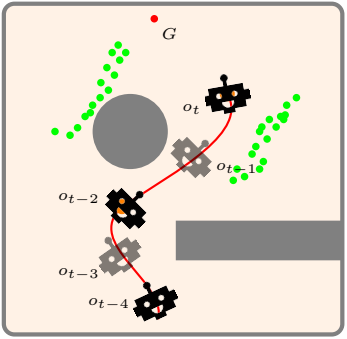

\setlength{\abovecaptionskip}{-0.3em}
\centering
\begin{tikzpicture}
	\fill[rounded corners, orange, ultra thick, fill opacity=0.1] (-3.0, -0.2) rectangle (1.5, 4.2);
	\draw[rounded corners, gray, ultra thick] (-3.0, -0.2) rectangle (1.5, 4.2);
	\draw[red, thick] (-0.95, 0) .. controls (-0.82, 0.5) and (-1.96, 1) .. (-1.43, 1.5);
	\draw[red, thick] (-1.43, 1.5) .. controls (-0.65, 2) and (0.26, 2.5) .. (-0.03, 3);
	\fill[orange] (-1.43, 1.5) circle (.1cm);
	\fill[orange] (-0.03, 3) circle (.1cm);
	\draw (-1.48,0.88) node[opacity=0.5] {\includegraphics[scale=0.04, angle=35]{Fig/robot.png}};
	\draw (-1,0.26) node {\includegraphics[scale=0.04, angle=25]{Fig/robot.png}};
	\draw (-1.35, 1.5) node {\includegraphics[scale=0.04, angle=-45]{Fig/robot.png}};
	\draw (-0.03, 3) node {\includegraphics[scale=0.04, angle=10]{Fig/robot.png}};
	\draw (-0.48,2.18) node[opacity=0.5] {\includegraphics[scale=0.04, angle=-45]{Fig/robot.png}};
	
	\fill[gray, ultra thick] (-0.7, 0.8) rectangle (1.5, 1.3);
	\draw[gray, thick] (-0.7, 0.8) rectangle (1.5, 1.3);
	
	\fill[gray] (-1.32, 2.5) circle (.5cm);
	
	\fill[green] (-2.32, 2.5) circle (.05cm);
	\fill[green] (-2.12, 2.45) circle (.05cm);
	\fill[green] (-2.02, 2.55) circle (.05cm);
	\fill[green] (-1.92, 2.7) circle (.05cm);
	\fill[green] (-1.85, 2.75) circle (.05cm);
	\fill[green] (-1.82, 2.85) circle (.05cm);
	\fill[green] (-1.72, 2.95) circle (.05cm);
	\fill[green] (-1.61, 3.05) circle (.05cm);
	\fill[green] (-1.53, 3.25) circle (.05cm);
	\fill[green] (-1.46, 3.45) circle (.05cm);
	\fill[green] (-1.38, 3.55) circle (.05cm);
	\fill[green] (-1.71, 3.15) circle (.05cm);
	\fill[green] (-1.63, 3.35) circle (.05cm);
	\fill[green] (-1.56, 3.55) circle (.05cm);
	\fill[green] (-1.48, 3.65) circle (.05cm);
	
	\fill[green] (0.05, 1.85) circle (.05cm);
	\fill[green] (0.1, 2) circle (.05cm);
	\fill[green] (0.2, 1.9) circle (.05cm);
	\fill[green] (0.3, 2.2) circle (.05cm);
	\fill[green] (0.4, 2.0) circle (.05cm);
	\fill[green] (0.35, 2.3) circle (.05cm);
	\fill[green] (0.45, 2.1) circle (.05cm);
	\fill[green] (0.4, 2.5) circle (.05cm);
	\fill[green] (0.5, 2.4) circle (.05cm);
	\fill[green] (0.43, 2.56) circle (.05cm);
	\fill[green] (0.62, 2.56) circle (.05cm);
	\fill[green] (0.53, 2.66) circle (.05cm);
	\fill[green] (0.72, 2.66) circle (.05cm);
	\fill[green] (0.68, 2.70) circle (.05cm);
	\fill[green] (0.74, 2.72) circle (.05cm);
	\fill[green] (0.76, 2.85) circle (.05cm);
	\fill[green] (0.89, 2.95) circle (.05cm);
	
	\node[ultra thick, font=\fontsize {6} {6}\selectfont] at (-1.6,0.2) {$o_{t-4}$};
	\node[ultra thick, font=\fontsize {6} {6}\selectfont] at (-2.0,0.6) {$o_{t-3}$};
	\node[ultra thick, font=\fontsize {6} {6}\selectfont] at (-2.0,1.6) {$o_{t-2}$};
	\node[ultra thick, font=\fontsize {6} {6}\selectfont] at (0.1,2.0) {$o_{t-1}$};
	\node[ultra thick, font=\fontsize {6} {6}\selectfont] at (-0.5,2.8) {$o_t$};
	
	\fill[red] (-1.0, 4.0) circle (.05cm);
	\node[ultra thick, font=\fontsize {6} {6}\selectfont] at (-0.8,3.8) {$G$};
	
	\end{tikzpicture}
\caption{Consecutive path-based frames for the DRL-based navigation policy. The gray circular and rectangular block denote static obstacles, green dots denote the trajectories of pedestrians, and $G$ denotes the target point. In the current scenario, the three consecutive path-based frames are expressed as $\Vec{\boldsymbol{s}}_t=(s_{t-4}, s_{t-2}, s_t)$, as the robot takes 2 time steps to follow each of the two previous paths, and sensor-based frames are expressed as $\Vec{\boldsymbol{s}}_t=(s_{t-2}, s_{t-1}, s_t)$.}
\label{fig:multi_frames}
\vspace{-1.0em}
\end{figure}

As shown in Fig.~\ref{fig:multi_frames}, we can specify the input of the DRL policy, i.e., three consecutive path-based frames, as $\Vec{\boldsymbol{s}}_t=(s_{t-(\tau_1+\tau_2)}, s_{t-\tau_1}, s_t)$, where  $s_{t-(\tau_1+\tau_2)}=(m_t^{-2}, g_t^{-2})$ denotes an egocentric costmap and a relative target pose w.r.t. the pose $(x_t^{-2}, y_t^{-2}, \alpha_t^{-2})$ of the robot, $s_{t-\tau_1}=(m_t^{-1}, g_t^{-1})$ w.r.t. the pose $(x_t^{-1}, y_t^{-1}, \alpha_t^{-1})$, and $s_t = (m_t, g_t)$ w.r.t. the pose $(x_t, y_t, \alpha_t)$. 
We use $\tau_1$ to denote the amount of the time steps for following the path $P^{-1}_t$ and $\tau_2$ to denote the amount of the time steps for following the path $P^{-2}_t$.


\subsubsection{Action Space} \label{method: action}
In this paper, we intend to train a DRL policy that outputs navigation paths directly.
In our formulation, we use B\'{e}zier curve to fit the navigation path via the current position of the robot and other $n$ control points generated by the DRL policy.
As shown in Fig.~\ref{fig:first}, the action space is the possible control points that can be selected in the rectangular area in front of the robot, which is high-dimensional.
Therefore, to reduce the dimension of the action space and the training difficulty, we discretize the selection of longitudinal coordinate values for these $n$ control points.

In specific, the $n$ control points are identified as $(l_t^1, h_t^1)$, $(l_t^2, h_t^2)$, $\ldots$, $(l_t^{n}, h_t^{n})$, where the point $(l, h)$ denotes the local longitudinal coordinate $l$ and the local horizontal coordinate $h$ when the current pose $(x_t, y_t, \alpha_t)$ of the robot is considered as the origin of the local coordinates. 
We assign a fixed interval between two adjacent longitudinal coordinates, i.e., $l_t^{i+1} - l_t^i =  l_t^1 =d$ for $1\leq i < n$.
In our experiments, we set $n=3$ and $d= 0.4/3 m$, then $l_t^1$, $l_t^2$, and $l_t^3$  are assigned to be $0.4/3m$, $0.8/3m$, and $0.4m$, respectively.



Notice that, the previous path $P_t^{-1}$ is specified by the starting pose $(x_{t-\tau}, y_{t-\tau}, \alpha_{t-\tau})$ of the robot and $n$ local control points $(l_{t-\tau}^1, h_{t-\tau}^1)$, $\ldots$, $(l_{t -\tau}^n, h_{t-\tau}^n)$, where $\tau$ is the amount of the time steps for following the path $P_t^{-1}$. 
Particularly, the current position $(x_t, y_t)$ of the robot at time~$t$ is the ending position of the robot for the previous path $P_t^{-1}$, then $(x_t, y_t)$ is equivalent to the global position for the last control point $(l_{t -\tau}^n, h_{t-\tau}^n)$ of $P_t^{-1}$, i.e.,
\begin{align*}
x_t & = l_{t -\tau}^n \cos(\alpha_{t-\tau}) - h_{t-\tau}^n \sin(\alpha_{t-\tau}) + x_{t-\tau},\\
y_t & = l_{t -\tau}^n \sin(\alpha_{t-\tau}) + h_{t-\tau}^n \cos(\alpha_{t-\tau}) + y_{t-\tau}.
\end{align*}



\subsubsection{Reward Function}
The DRL policy aims to maximize the cumulative reward. 
In robot navigation tasks, the primary goal of the robot is to reach the target without collision within a limited time. 
Based on this goal, the robot's trajectory is further required to be smoother. 
Therefore, the reward function of the algorithm is defined as follows:
\[
    r_t = r_t^\textit{goal} + r_t^\textit{safe} + r_t^\textit{curvature} + r_t^\textit{straight},
\]
where $r_t$ is the sum of four parts, $r_t^\textit{goal}$, $r_t^\textit{safe}$, $r_t^\textit{curvature}$ and $r_t^\textit{straight}$.

In particular, $r_t^\textit{goal}$ is the reward when the robot reaches the local target:
\[
r_t^{\textit{goal}} = \begin{cases}
r_{arr}, & \text{if target is reached,} \\
0, & \text{otherwise.} 
\end{cases}
\]

$r_t^{\textit{safe}}$ specifies the penalty when the robot encounters a collision:
\[
r_t^\textit{safe} = \begin{cases}
r_{col}, & \text{if collision,} \\
0, & \text{otherwise.} 
\end{cases}
\]

We use the $r_t^\textit{curvature}$ to encourage the DRL algorithm to output paths with less curvature:
\[
r_t^\textit{curvature} = - \varepsilon_{1}(\sum_{i=0}^{N} k(x)), x = \frac{i}{N} \in [0, 1],
\]
where $N$ is the total number of points after the discretization of the path, $\varepsilon_{1}$ is a hyper-parameter and $k(x)$ is the curvature of the path at the $i$th point. 
The following formula defines $k(x)$:
\[
k(x) = \frac{\Vert B^{'}(x) \times B^{''}(x) \Vert}{{\Vert B^{'}(x) \Vert}^3},
\]
where $B(x)$ means Quadratic B\'{e}zier curve function.
The intuition behind the definition of $r_t^\textit{curvature}$ is that the larger the curvature of points along the path, the more unstable the robot's motion becomes, and thus the penalty should be greater.

The purpose of $r_t^\textit{straight}$ is to encourage the DRL algorithm to generate paths that maximize alignment with the direction of the vehicle body. This strategic approach ensures the smooth transition between adjacent paths, thereby enhancing the smoothness of the robot's movement trajectory.
\[
r_t^\textit{straight} = \varepsilon_{2}(L_{min} - L),
\]
where $\varepsilon_{2}$ is a hyper-parameter, $L_{min}$ is the length of the shortest path, which is related to the action space dimension and the distance between the path points, and $L$ is the length of the path output by the DRL algorithm. 
The intuition behind the definition of $r_t^\textit{straight}$ is to avoid sharp angles between adjacent paths.
In our experiments, we set $r_{arr} = 500$, $r_{col} = -700$, $N = 100$, $\varepsilon_{1} = 0.4$ and $\varepsilon_{2} = 200$.

\IncMargin{1em}
\begin{algorithm}[htp]
	 \emph{Initialize the policy network $\pi$, the value network, and various hyperparameters}\; 
      \emph{Clear the experience buffer Buffer $\mathcal{D}$}\;
	 \For{epoch = $1, \ldots, E$}{
      \For{step t = $1, \ldots, T_{ep}$}{
        $a_t = \pi(\Vec{\boldsymbol{s}}_t)$\;
        $\Vec{\boldsymbol{a}}_t = B\acute{e}zier(a_t)$\;
        \If{A path has been followed in its entirety}{
        $a = \Vec{\boldsymbol{a}}_t$\;
        }
        $s_{t+1}, r_t = path\_follow(a)$\; 
        \If{A path has been followed in its entirety}{
        \emph{deque $\Vec{\boldsymbol{s}}_{t+1}$ append $s_{t+1}$}\;
        $\mathcal{D} \leftarrow \mathcal{D} \cup \{\Vec{\boldsymbol{s}}_t, a_t, r_t, \Vec{\boldsymbol{s}}_{t+1}\}$\;
        }
        \emph{Sample batch $\mathcal{B} \sim \mathcal{D}$ to update SAC network}\;
        \If{robot has stopped}{
        $\Vec{\boldsymbol{s}}_t = reset()$\;
        }
      }
      } 
 	 \caption{SAC for the generation of paths}
 	 \label{SAC_code} 
\end{algorithm}
\vspace{-1.0em}
\DecMargin{1em}

\begin{figure}[h]
    \vspace{-1.0em}
    \setlength{\abovecaptionskip}{-0.3em}
    \centering
    \includegraphics[width=0.5\textwidth]{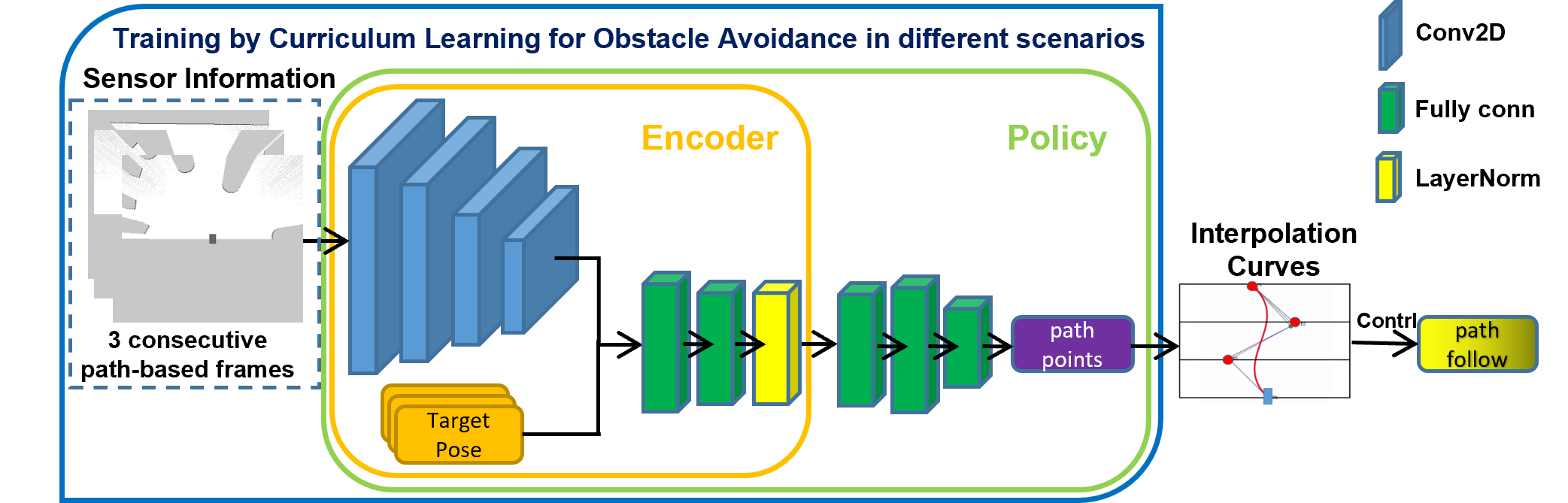}
    \caption{An overview of the entire process and the structure of the policy network. The egocentric costmap is fed into a fully convolutional network for the feature extraction. The extracted features are then concatenated with the relative target pose of the robot and input into a fully connected network to output the final coordinates of the path points.}
    \label{fig:overview}
    \vspace{-1.0em}
\end{figure}

\subsection{Training by Curriculum Learning for Obstacle Avoidance}


It remains challenging to achieve an optimal policy through the above method stably and efficiently.
Therefore, we employ curriculum learning~\cite{bengio2009curriculum} to assist and speed up the training process. 
Overall, we design a two-stage training process for curriculum learning in our experiments. 
In the first stage, we perform the DRL agent in two kinds of environments for the training,
both of which were conducted in a static scenario with gradually increased complexity (4 and 16 static obstacles, respectively). 
At this stage, we intent to enable the DRL agent to learn basic navigation abilities and obstacle avoidance capabilities.
In the second stage, we continuously train the DRL agent obtained from the first stage in environments with much more complex dynamic scenarios, which enable the DRL agent to learn a more robust navigation strategy to interact with dynamic obstacles. 


\section{Experiments}
\subsection{Experiments on simulation scenarios}



\begin{figure}[h]
    \vspace{-0.5em}
    \setlength{\abovecaptionskip}{-0.3em}
    \centering
    \includegraphics[width=0.45\textwidth]{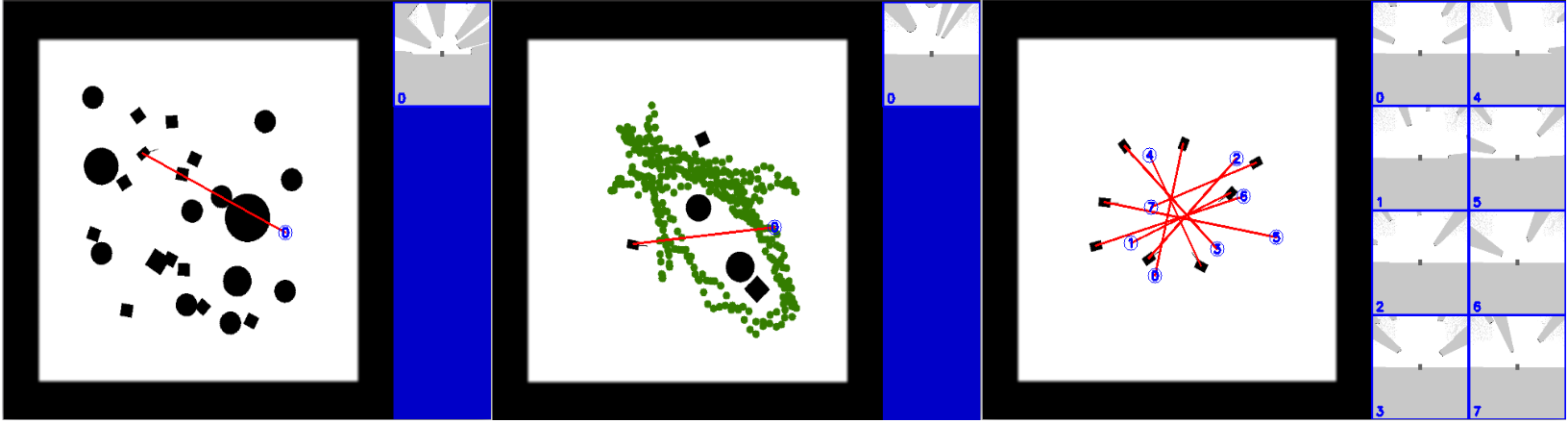}
    \caption{Illustration of the three scenarios for the training. The blue digital circles indicate the target position of the corresponding robots, red lines represent the straight paths from the starting point to the target point for robots, black circle and rectangle blocks specify static obstacles, green dots denote pedestrian trajectories, and blue boxes on the right show sensor maps of each robot.}
    \label{fig:succ_cur_vel}
    \vspace{-1.0em}
\end{figure}

Here we consider three different types of scenarios, i.e., static, pedestrian, and multi-agent scenarios, to generate random environments in our customized simulator~\cite{chen2020distributed} to collect experiences for the training of the navigation policy.
Fig.~\ref{fig:succ_cur_vel} illustrates environments for these three scenarios, respectively.

We compare the performance of \textit{PathRL} with three other DRL-based navigation methods, i.e., PPO~\cite{chen2020distributed}, SAC~\cite{haarnoja2018soft} and PSD~\cite{yao2021crowd}.
All methods are evaluate in both static and dynamic simulation environments using an Ackermann steering robot\footnote{Note that, it is more challenging for DRL methods to train a navigation policy for a robot with additional kinematic constraints like an Ackermann steering robot comparing with a differential robot. We use B\'{e}zier curves to guarantee kinematic feasibility. Dynamic constraints (acceleration, curvature) are satisfied by forcibly limiting planning parameters within reasonable ranges.}
whose length by width is $0.3\times 0.2$. 
The action space for both PPO and SAC methods is the linear velocity and the angle of the front wheel, i.e., $v \in [0, 0.6]$ and $\theta \in [-0.785, 0.785]$. 
Notice that, both PPO and SAC methods share the similar network structure as \textit{PathRL} as shown in Fig.~\ref{fig:overview}. 
PSD is a crowd navigation method that enables efficient and smooth motion planning within highly crowded and dynamic environments.
It's important to mention that we have modified the network output of PSD to support Ackermann steering robots.
We have also well justified the training details for all methods to achieve good performance as shown in our previous work~\cite{chen2020distributed, qiu2022learning}.
In this paper, we show that DRL policies as trained by \textit{PathRL} that directly output paths can greatly improve the speed stability and trajectory smoothness of the robot.

We introduce four metrics to evaluate the performance of navigation policies trained by different DRL methods in multiple scenarios as the following: 
\begin{itemize}
    \item \textit{Success rate (SUCC)}: the radio of episodes in which the robot reaches the target pose without collision.
    \item \textit{Average curvature (CUR)}: the average curvature of the robot's driving path at each episode, which evaluates the stability of the (angular) speeds and the smoothness of the trajectories for the robot. 
    \item \textit{Average length (LEN)}: the average length of the robot's driving path at each success episode.
    \item \textit{Average time (TIME)}: the average cost time at each success episode.
\end{itemize}

In the following, all experimental results for different methods were obtained from the average results for 1000 randomly generated environments of each scenario.
Pedestrians in the scenario are driven by OCRA and SFM and multi-agents adopt the self-play strategy without communicating with each other as shown in~\cite{chen2020distributed}.
For a fair comparison, we trained 3 versions of the models for each DRL method and summary the average performance of each method for multiple scenarios in following tables.

\subsubsection{Comparative experiments}

\begin{table}[htb]
        \vspace{-1.0em}
	\setlength{\abovecaptionskip}{-1em}
	\centering
	\caption{Performance of Different Navigation Methods}
	\label{tab: comparative_exp}
	\newcommand{\tabincell}[2]{\begin{tabular}[t]{@{}#1@{}}#2\end{tabular}}
        \resizebox{\columnwidth}{!}{
	\begin{tabular}[t]{c|c|cccc}
		\bottomrule[2pt]
		Scenarios &Methods   &SUCC(\%)$\uparrow$   &CUR$\downarrow$  &LEN(m)$\downarrow$ &TIME(s)$\downarrow$ \\
            \noalign{\hrule height 1pt}
		\multirow{4}*{\shortstack{Static Scenario \\ (16 Obstacles)}} & 
		\tabincell{c}{PPO}    & \tabincell{c}{0.979} & \tabincell{c}{1.237}   & \tabincell{c}{5.209}  & \tabincell{c}{9.177} 
		\\
		& \tabincell{c}{SAC}    & \tabincell{c}{0.972} & \tabincell{c}{1.418}   & \tabincell{c}{5.486}  & \tabincell{c}{9.474} 
		\\
        & \tabincell{c}{PSD}    & \tabincell{c}{0.975} & \tabincell{c}{1.219}   & \tabincell{c}{5.136}  & \tabincell{c}{9.164} 		\\
		& \tabincell{c}{PathRL}    & \tabincell{c}{\textbf{0.985}} & \tabincell{c}{\textbf{0.691}}   & \tabincell{c}{\textbf{4.804}}  & \tabincell{c}{\textbf{7.522}} 
		\\
		\hline
            \multirow{4}*{\shortstack{Static Scenario \\ (24 Obstacles)}} &
		\tabincell{c}{PPO}    & \tabincell{c}{0.960} & \tabincell{c}{1.454}   & \tabincell{c}{5.292}  & \tabincell{c}{9.113} 
		\\
		& \tabincell{c}{SAC}    & \tabincell{c}{0.941} & \tabincell{c}{1.571}   & \tabincell{c}{5.761}  & \tabincell{c}{10.004} 
		\\
        & \tabincell{c}{PSD}    & \tabincell{c}{0.958} & \tabincell{c}{1.405}   & \tabincell{c}{5.307}  & \tabincell{c}{9.154} 
		\\
		& \tabincell{c}{PathRL}    & \tabincell{c}{\textbf{0.968}} & \tabincell{c}{\textbf{0.772}}   & \tabincell{c}{\textbf{5.281}}  & \tabincell{c}{\textbf{8.249}} 
		\\
		\hline
            \multirow{4}*{\shortstack{Dynamic Scenario \\ (4 Obstacles \& 4 Pedestrians)}} &
		\tabincell{c}{PPO}    & \tabincell{c}{0.850} & \tabincell{c}{2.519}   & \tabincell{c}{6.921}  & \tabincell{c}{11.850} 
		\\
		& \tabincell{c}{SAC}    & \tabincell{c}{0.889} & \tabincell{c}{1.538}   & \tabincell{c}{\textbf{6.858}}  & \tabincell{c}{\textbf{11.533}} 
		\\
        & \tabincell{c}{PSD}    & \tabincell{c}{\textbf{0.911}} & \tabincell{c}{1.584}   & \tabincell{c}{7.483}  & \tabincell{c}{12.988} 
		\\
		& \tabincell{c}{PathRL}    & \tabincell{c}{0.889} & \tabincell{c}{\textbf{0.904}}   & \tabincell{c}{8.431}  & \tabincell{c}{12.632} 
		\\
		\hline
            \multirow{4}*{\shortstack{Dynamic Scenario \\ (6 Pedestrians)}} &
		\tabincell{c}{PPO}    & \tabincell{c}{0.769} & \tabincell{c}{2.027}   & \tabincell{c}{\textbf{7.587}}  & \tabincell{c}{\textbf{14.103}}
		\\
		& \tabincell{c}{SAC}    & \tabincell{c}{0.894} & \tabincell{c}{1.462}   & \tabincell{c}{8.452}  & \tabincell{c}{14.224} 
		\\
        & \tabincell{c}{PSD}    & \tabincell{c}{\textbf{0.926}} & \tabincell{c}{1.387}   & \tabincell{c}{9.094}  & \tabincell{c}{15.867} 
		\\
		& \tabincell{c}{PathRL}    & \tabincell{c}{0.875} & \tabincell{c}{\textbf{0.836}}   & \tabincell{c}{14.41}  & \tabincell{c}{20.268} 
		\\
		\hline
            \multirow{4}*{\shortstack{Dynamic Scenario \\ (8 Agents)}} &
		  \tabincell{c}{PPO}    & \tabincell{c}{0.879} & \tabincell{c}{2.197}   & \tabincell{c}{\textbf{6.171}}  & \tabincell{c}{11.832} 
		\\
		& \tabincell{c}{SAC}    & \tabincell{c}{0.947} & \tabincell{c}{3.869}   & \tabincell{c}{7.004}  & \tabincell{c}{12.702} 
		\\
        & \tabincell{c}{PSD}    & \tabincell{c}{0.909} & \tabincell{c}{1.928}   & \tabincell{c}{7.409}  & \tabincell{c}{12.364} 
		\\
		& \tabincell{c}{PathRL}    & \tabincell{c}{\textbf{0.998}} & \tabincell{c}{\textbf{0.705}}   & \tabincell{c}{7.567}  & \tabincell{c}{\textbf{11.642}} 
		\\
		\toprule[2pt]
	\end{tabular}}
        \vspace{-1.0em}
\end{table}

Table~\ref{tab: comparative_exp} shows that \textit{PathRL} outperforms the other DRL methods by a considerable improvement of ``Average curvature (CUR)'' without sacrificing ``Success rate (SUCC)''. 
In dynamic environments, we find out that both ``Average length (LEN)'' and ``Average time (TIME)'' of \textit{PathRL} are slightly larger than that of other DRL methods, as \textit{PathRL} is required to drive the robot to avoid dynamic obstacles stably and smoothly, which is preferred for most real-world applications. Notice that, compared to \textit{PathRL}, PSD has additional pedestrian map input, resulting in a higher success rate in pedestrian scenarios. 

Notice that, additional smoothing requirements are preferred for an Ackermann steering robot.
We use the average changes of the steering angle at each step, i.e. $\Delta\theta = \theta_{t+1} - \theta_{t}$, as a metric to assess the smoothness and comfort of the robot's movement.

\begin{table}[htb]
    \vspace{-1.0em}
    \setlength{\abovecaptionskip}{-1em}
    \centering
    \caption{Average Changes of Steering Angle for Different Methods}
    \newcommand{\tabincell}[2]{\begin{tabular}[t]{@{}#1@{}}#2\end{tabular}}
    \label{tab:steering}
    \resizebox{0.8\columnwidth}{!}{
    \begin{tabular}[t]{c|c|c}
    \bottomrule[2pt]
       Scenarios  & Methods & $\Delta\theta(rad)\downarrow$ \\
    \noalign{\hrule height 1pt}
	\multirow{4}*{\shortstack{Static Scenario \\ (24 Obstacles)}} &
        \tabincell{c}{PPO}    & \tabincell{c}{0.5527} \\
        & \tabincell{c}{SAC}    & \tabincell{c}{0.5584} \\
        & \tabincell{c}{PSD}    & \tabincell{c}{0.5498} \\
        & \tabincell{c}{PathRL}    & \tabincell{c}{\textbf{0.0093}} \\
    \hline
	\multirow{4}*{\shortstack{Dynamic Scenario \\ (4 Obstacles \& 4 Pedestrians)}} &
        \tabincell{c}{PPO}    & \tabincell{c}{0.8983} \\
        & \tabincell{c}{SAC}    & \tabincell{c}{0.5047} \\
        & \tabincell{c}{PSD}    & \tabincell{c}{0.4763} \\
        & \tabincell{c}{PathRL}    & \tabincell{c}{\textbf{0.0114}} \\
    \hline
	\multirow{4}*{\shortstack{Dynamic Scenario \\ (8 Agents)}} &
        \tabincell{c}{PPO}    & \tabincell{c}{1.2490} \\
        & \tabincell{c}{SAC}    & \tabincell{c}{1.1156} \\
        & \tabincell{c}{PSD}    & \tabincell{c}{1.0271} \\
        & \tabincell{c}{PathRL}    & \tabincell{c}{\textbf{0.0023}} \\
    \toprule[2pt]
    \end{tabular}}
    \vspace{-2.0em}
\end{table}

Table~\ref{tab:steering} summarizes the results of the $\Delta\theta$ metric for three reinforcement learning methods. Since \textit{PathRL} directly outputs the path and decouples the motion control module from the DRL policy, it achieves a dramatic improvement of the stability of the steering angle for the robot comparing with other DRL methods. 


In our setting, the DRL navigation policies are trained using the experiences collected in various simulation environments. Then it is crucial for these policies to be robust in the presence of the differences between the simulation model and the real-world robot platform.
Notice that, DRL-based navigation methods that directly commands the robot with low-level controls are sensitive to parameters of the robot platform, which limits the generality of methods.
Meanwhile, \textit{PathRL} directly outputs the path and decouples the motion control module from the DRL policy. 
Then, sophisticated velocity-planning and path-following controllers, whose parameters are specified according to the real-world robot platform, can be applied for the robot to follow the paths generated by the DRL policy.
In the following, we consider the impact for the navigation performance for different DRL-based navigation methods when varying the maximum driving speed of the robot platform.

\begin{figure}[ht]
    \vspace{-1.5em}
    \setlength{\abovecaptionskip}{-1.0em}
    \centering
    \includegraphics[width=0.35\textwidth]{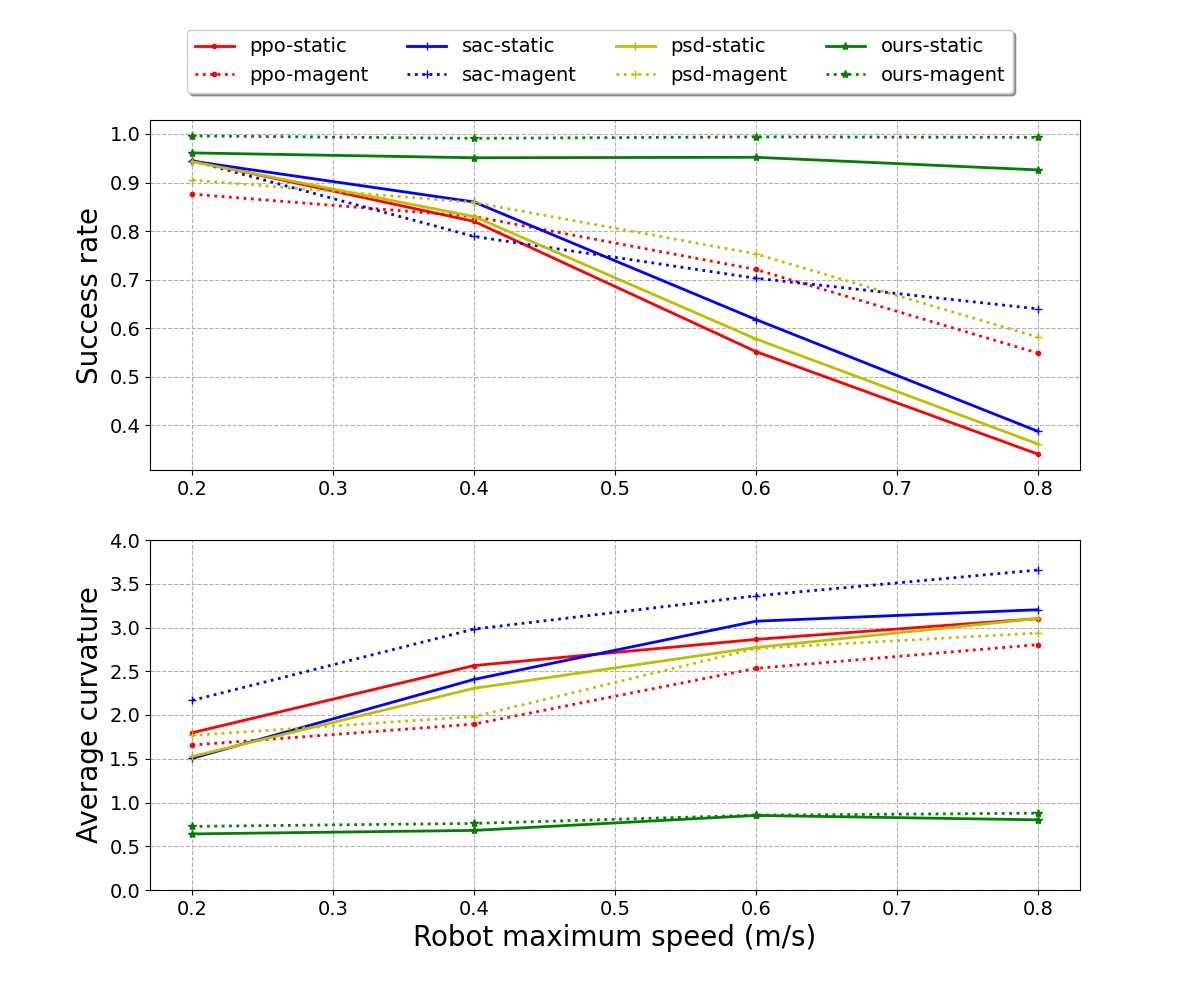}
    \caption{The performance of DRL methods after increasing the maximum moving speed of the robot in different scenarios.}
    \label{fig:succ_cur_vel1}
    \vspace{-1.0em}
\end{figure}

In our experiment, we respectively apply the four DRL methods to train corresponding navigation policies in the same simulation environments, where the maximum moving speed of the robot is 0.2 m/s. 
Then we test the performance of these navigation policies in both static and multi-agent scenarios when the maximum moving speed is increased to 0.4 m/s, 0.6 m/s, and 0.8 m/s, respectively.
The experimental results are summarized in Figure~\ref{fig:succ_cur_vel1}, where ``x-static'' and ``x-magent' denote the performance of the trained navigation policy using the DRL method ``x'' tested in static and mullti-agent scenarios, respectively.
Comparing with other DRL methods, the performance of \textit{PathRL} is more robust when the maximum moving speed is increased, which maintains the stable success rate and average curvature of moving trajectories.

\begin{figure}[h]
    \setlength{\abovecaptionskip}{-0.3em}
    \centering
    \includegraphics[width=0.35\textwidth]{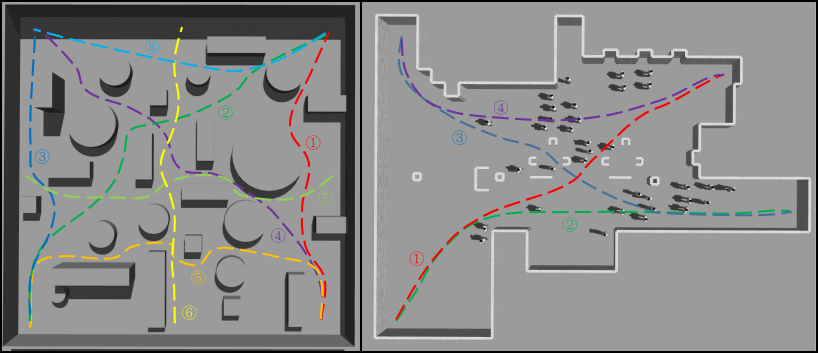}
    \caption{3D Simulation test environments and test routes.}
    \label{fig:gazebo}
    \vspace{-1.0em}
\end{figure}

Additionally, we compare the performance of \textit{PathRL} with TEB algorithm in 3D gazebo environments.
Notice that, the DRL navigation policy is trained in previous training environments using \textit{PathRL}, which is only tested in gazebo environments here and We make every effort to fine-tune the parameters of the TEB algorithm to achieve improved performance.
We test \textit{PathRL} and TEB in environments of two scenarios, i.e., static scenario and dynamic scenario. 
As illustrated in Fig.~\ref{fig:gazebo}, the testing environments of the static scenario is equipped  with 8 possible routes (i.e., global paths) with different origins and destinations surrounding with static obstacles.
The testing environments of the dynamic scenario contains 4 different routes surrounding with 25--35 pedestrians which are guided by the social forces model~\cite{helbing1995social, moussaid2009experimental, xie2023drl}. 
Table~\ref{tab: tra} summarizes the performance of \textit{PathRL} and TEB in testing gazebo environments, where each method is tested in the same environment for five times. 
In addition, we use ``NCOLL'' to denote the average number of collisions during the execution of the policy or algorithm in testing environments.


\begin{table}[htb]
	\setlength{\abovecaptionskip}{-1.5em}  
	\centering
	\caption{Comparing \textit{PathRL} and TEB in 3D gazebo environments}
	\label{tab: tra}
	\newcommand{\tabincell}[2]{\begin{tabular}[t]{@{}#1@{}}#2\end{tabular}}

	\renewcommand\arraystretch{1.0}
        \resizebox{0.8\columnwidth}{!}{
	\begin{tabular}[t]{c|c|cccc}
		\bottomrule[2pt]   
		Scenarios &Method   &SUCC(\%)$\uparrow$   &NCOLL$\downarrow$ & TIME(s)$\downarrow$\\
		\noalign{\hrule height 1pt}
            \multirow{2}*{Static Scenario}
		& \tabincell{c}{PathRL}    & \tabincell{c}{\textbf{0.900}} & \tabincell{c}{\textbf{3}}   & \tabincell{c}{\textbf{65.326}}
		\\
		& \tabincell{c}{TEB}    & \tabincell{c}{0.825} & \tabincell{c}{5}   & \tabincell{c}{77.368}
		\\
		\hline
            \multirow{2}*{Dynamic Scenario}
		& \tabincell{c}{PathRL}    & \tabincell{c}{\textbf{0.850}} & \tabincell{c}{\textbf{5}}   & \tabincell{c}{\textbf{76.331}}  
		\\
		& \tabincell{c}{TEB}    & \tabincell{c}{0.750} & \tabincell{c}{8}   & \tabincell{c}{79.606}  
		\\
		\toprule[2pt]
	\end{tabular}}
        \vspace{-1.0em}
\end{table}

\subsubsection{Ablation study}

The input of DRL navigation position consists three consecutive state frames.
In \textit{PathRL}, there are two alternative considerations of ``consecutive'', i.e., 
$\romannumeral1$) sensor-based three consecutive frames, where  $\Vec{\boldsymbol{s}}_t=(s_{t-2}, s_{t-1}, s_t)$; $\romannumeral2$) path-based three consecutive frames, where $\Vec{\boldsymbol{s}}_t=(s_{t-(\tau_1+\tau_2)}, s_{t-\tau_1}, s_t)$.
Notice that, path-based consecutive frames have greater time intervals.
The path-based frames introduces information for a longer span of time, which can benefit robot collision avoidance in dynamic scenarios and ensure the smooth transition between adjacent paths.
To evaluate the advantage of path-based consecutive frames, 
we consider following ablation experiments. 
Specifically, we compare the performance of our design with sensor-based three consecutive frames in a dynamic scenario. 
As shown in Table~\ref{tab:state}, the design of path-based 
three consecutive frames achieves a higher success rate and smoother driving trajectory than the design of sensor-based three consecutive  frames, with slight larger values of ``LEN'' and ``TIME''.

\begin{table}[htb]
\vspace{-1.0em}
\setlength{\abovecaptionskip}{0cm}
\centering
\caption{Performance of Different State Input}
\label{tab:state}
\resizebox{\columnwidth}{!}{
\begin{tabular}{c|cccc}
\bottomrule[2pt]
Setting& SUCC(\%)$\uparrow$& CUR$\downarrow$& LEN(m)$\downarrow$& TIME(s)$\downarrow$\\
\noalign{\hrule height 1pt}
path-based frames& \textbf{0.889}& \textbf{0.993}& 8.758 & 12.632\\
sensor-based frames& 0.837 & 1.469 & \textbf{7.905} & \textbf{11.590} \\
\toprule[2pt]
\end{tabular}}
\vspace{-1.0em}
\end{table}

Our experiments show that the $r_{straight}$ term in the reward function plays an important role in improving the experimental results and guaranteeing the smooth transition. On the one hand, without the $r_{straight}$ constraint, the robot would twist during its movement, resulting in an uneven driving trajectory. On the other hand, the inclusion of the $r_{straight}$ term also results a significant improvement of the navigation performance. Table~\ref{tab:reward} summarizes the ablation study. 

\begin{table}[htb]
        \vspace{-1.0em}
	\setlength{\abovecaptionskip}{-1em}  
	\centering
	\caption{Performance of Different Reward Function Settings}
	\label{tab:reward}
	\newcommand{\tabincell}[2]{\begin{tabular}[t]{@{}#1@{}}#2\end{tabular}}
	\renewcommand\arraystretch{1.0}
        \resizebox{\columnwidth}{!}{
	\begin{tabular}[t]{c|c|cccc}
		\bottomrule[2pt]   
		Scenarios &Rewards   &SUCC(\%)$\uparrow$   &CUR$\downarrow$  &LEN(m)$\downarrow$ &TIME(s)$\downarrow$ \\
		\noalign{\hrule height 1pt}
            \multirow{2}*{\shortstack{Static Scenario \\ (24 Obstacles)}}
		& \tabincell{c}{PathRL}    & \tabincell{c}{\textbf{0.968}} & \tabincell{c}{\textbf{0.772}}   & \tabincell{c}{\textbf{5.281}}  & \tabincell{c}{\textbf{8.249}} 
		\\
		& \tabincell{c}{PathRL w.o. $r_{straight}$}    & \tabincell{c}{0.936} & \tabincell{c}{2.014}   & \tabincell{c}{7.227}  & \tabincell{c}{11.460} 
		\\
		\hline
            \multirow{2}*{\shortstack{Dynamic Scenario \\ (4 Obstacles \& 4 Pedestrians)}}
		& \tabincell{c}{PathRL}    & \tabincell{c}{\textbf{0.889}} & \tabincell{c}{\textbf{0.904}}   & \tabincell{c}{\textbf{8.758}}  & \tabincell{c}{\textbf{12.632}} 
		\\
		& \tabincell{c}{PathRL w.o. $r_{straight}$}    & \tabincell{c}{0.847} & \tabincell{c}{2.461}   & \tabincell{c}{9.408}  & \tabincell{c}{15.079} 
		\\
		\toprule[2pt]
	\end{tabular}}
        \vspace{-2.0em}
\end{table}

\subsection{Deploy to real-world Ackermann steering robot}

We deploy the trained DRL policy to a real-world Ackermann steering robot to test its performance in real-world static and dynamic environments for collision avoidance. 
We use Simultaneous Localization and Mapping (SLAM)~\cite{duan2022pfilter} technology for mapping and localization. 

\begin{figure}[ht]
    \setlength{\abovecaptionskip}{-0.3em}
    \centering
    \includegraphics[width=0.45\textwidth]{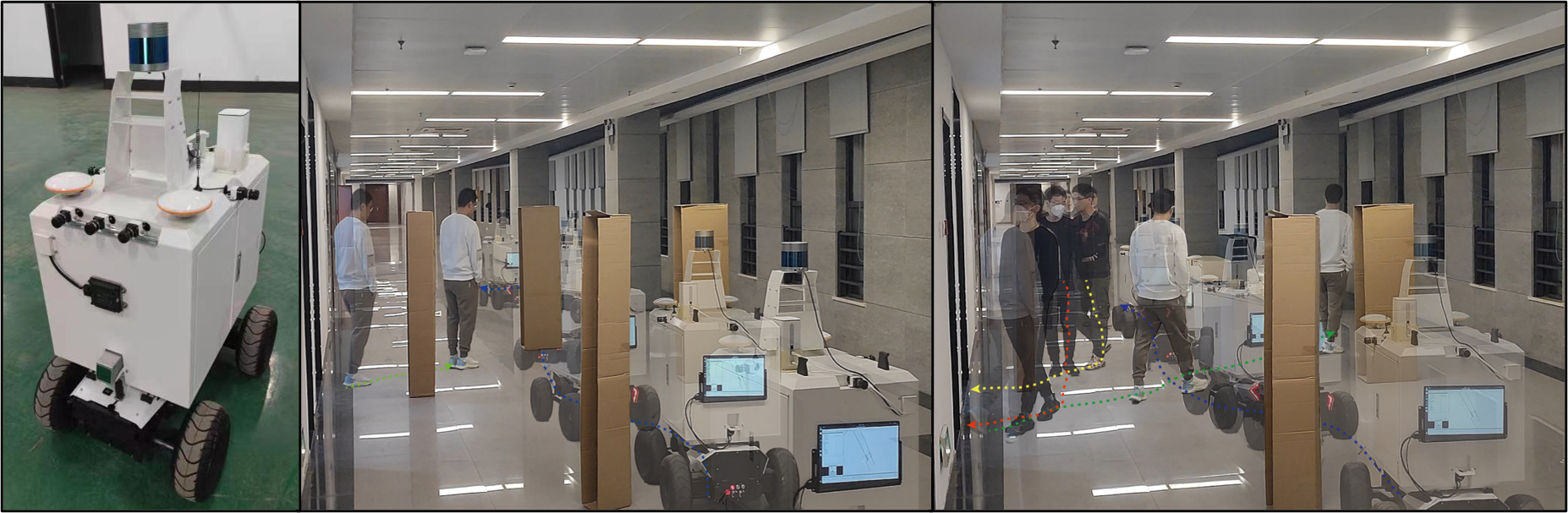}
    \caption{The robot and trajectories of the robot in the test environments. }
    \label{fig:real_env}
    \vspace{-1.0em}
\end{figure}

As shown in Fig.~\ref{fig:real_env}, the platform for the Ackermann steering robot is based on Agilex hunter2.0 chassis and uses a 32-line 3D laser sensor. In addition, the robot is equipped with an RTX 3090 as a computing unit. The size of the robot, length $\times$ width $\times$ height, is $0.95m \times 0.75m \times 1.45m$. Our experiments show that the robot can successfully avoid static and dynamic obstacles and complete navigation tasks. More illustrations of the performance of the robot are shown in our demonstration video.

\section{Conclusions}
In this paper, we present an end-to-end path generation method without any expert prior knowledge, \textit{PathRL}, for robot collision avoidance using DRL. 
We diminish the dimension of the policy space by decreasing the selection of longitudinal coordinate values for control points. By utilizing interpolation curves, we generate smoother paths. By employing three consecutive path-based frames and reward function constraints, we guarantee a smooth transition.
Additionally, we utilize curriculum learning to expedite the training process and acquire a more robust navigation policy.

Comparing with DRL-based navigation methods that directly output low-level control commands, \textit{PathRL} achieves fewer changes of driving trajectory curvature and less variation in angular rotation without sacrificing the success rate. 
Moreover, \textit{PathRL} is more robust for the differences between the simulation model and the real-world robot platform, as it provides the decoupling between the planning and the control modules.
In a variety of challenging scenarios in both simulation and the real world, \textit{PathRL} achieves the nice navigation performance.


\bibliographystyle{IEEEtran}
\bibliography{TRLNV}
\end{document}